\documentclass[conference]{IEEEtran}
\ifCLASSINFOpdf
  \usepackage[pdftex]{graphicx}
\else
\fi
\usepackage{todonotes}

\usepackage{algorithm}
\usepackage[noend]{algpseudocode}
\usepackage{algorithmicx}
\usepackage{hyperref}

\IEEEoverridecommandlockouts
\begin{document}
%
\title{QoS-Aware Multi-Armed Bandits\thanks{2016 IEEE. Personal use of this material is permitted.
Permission from IEEE must be obtained for all other uses, in any current or future 
media, including reprinting/republishing this material for advertising or promotional 
purposes, creating new collective works, for resale or redistribution to servers or 
lists, or reuse of any copyrighted component of this work in other works. 
DOI: \href{http://ieeexplore.ieee.org/document/7789452/}{10.1109/FAS-W.2016.36}}}

\author{
\IEEEauthorblockN{Lenz Belzner}
\IEEEauthorblockA{
Institute for Informatics\\
LMU Munich}
\and
\IEEEauthorblockN{Thomas Gabor}
\IEEEauthorblockA{
Institute for Informatics\\
LMU Munich}}


%


\maketitle

\begin{abstract}
Motivated by runtime verification of QoS requirements in self-adaptive and self-organizing systems that are able to reconfigure their structure and behavior in response to runtime data, we propose a QoS-aware variant of Thompson sampling for multi-armed bandits. It is applicable in settings where QoS satisfaction of an arm has to be ensured with high confidence efficiently, rather than finding the optimal arm while minimizing regret.
Preliminary experimental results encourage further research in the field of QoS-aware decision making.
\end{abstract}


%
\IEEEpeerreviewmaketitle

\section{Introduction}
We consider the problem of exploration and exploitation under uncertainty and QoS requirements.
Imagine a smart factory control system that is able to provide potential reconfigurations in response to events of change, e.g. on failure detection of a particular machine. In quality critical settings, such a situation may result in downtime until QoS requirements have been reestablished. For example, a factory is required to create products with a guaranteed maximum error rate. At the same time, the confidence about this error rate should be built as fast as possible.

One way to enable verification of QoS requirements at runtime is by performing statistical model checking of the system by using a simulation of the system and its application domain \cite{lee2007statistical}. Here, i.i.d. Monte Carlo simulations of system execution are performed until satisfaction or violation of a particular requirement has been proven up to a given confidence bound.

Given a set of potential reconfigurations in a new situation, QoS-aware automated runtime verification pursues two goals.
\begin{enumerate}
	\item Identify a configuration satisfying QoS requirements.
	\item Maximize the confidence about this configuration.
\end{enumerate}

This problem yields an exploration vs. exploitation tradeoff: Once a promising configuration has been identified, confidence about its quality should be maximized. However, the system is also interested in configurations with higher quality than the current promising candidate (because QoS confidence can be established faster for these configurations).

Multi-armed bandits (MAB) provide a well-studied formal framework for studying exploration vs. exploitation in decision making \cite{bubeck2012regret}. In this paper, we will outline an approach to QoS-aware decision making in the MAB framework. In Section \ref{sec:MAB}, we will formally describe the MAB framework and Thompson sampling, a baseline for MAB decision making based on Bayesian statistics. In Section \ref{sec:QATS}, we will describe how to perform QoS-aware Thompson sampling.

\section{Multi Armed Bandits}
\label{sec:MAB}

A multi-armed bandit is a set of distributions (e.g. of quality, payoff, utility, etc.). For simplicity, we restrict ourselves to Bernoulli distributions, which return a value of one with a probability of $p$, and zero otherwise.

A typical task is to identify the optimal arm while maximizing its payoff at the same time. In the case of Bernoulli bandits, the optimal arm $i$ is the one with maximal $p_i$. A state-of-the-art baseline approach to the bandit problem is Thompson sampling \cite{chapelle2011empirical,agrawal2012analysis}. It builds a distribution about possible values $p_i$ for each arm $i$, representing the decision makers uncertainty (or beliefs) about the distribution parameter based on its observations.

For Bernoulli bandits, a convenient choice for modeling parameter uncertainty is the Beta distribution with parameters $\alpha$ and $\beta$ \cite{press1989bayesian}. It is the conjugate prior of the Bernoulli distribution, allowing for efficient posterior computation and analysis \cite{press1989bayesian}. Given an arm $i$ with $s_i$ successes (i.e. $s_i$ times reward one was observed) and $f_i$ failures, and assuming a uniform distribution as prior, the posterior distribution about $p_i$ is given by $\textit{Beta}(s_i + 1, f_i + 1)$.

Thompson sampling is outlined in Algorithm \ref{alg:TS}. First, potential values for $p_i$ of each arm are sampled from the current belief distributions. Then, the arm with the best sample is played, and its observed outcomes are updated, effectively changing the belief about its parameter.
\begin{algorithm}[H]
	\begin{algorithmic}[1]
		\Procedure {Thompson Sampling}{}
		\State Sample $\hat{p}_i$ from $\textit{Beta}(s_i + 1, f_i + 1)$ for each arm $i$
		\State Play arm $i$ with $\max \hat{p}_i$
		\State Update $s_i$ or $f_i$ according to result
		\EndProcedure
	\end{algorithmic}
	\caption{Thompson sampling for Bernoulli bandits.}
	\label{alg:TS}
\end{algorithm}

Despite its simplicity, Thompson sampling has recently attracted research interest due to its theoretical properties and empirical success, showing comparable performance to other state-of-the-art bandit approaches such as UCB \cite{chapelle2011empirical,agrawal2012analysis}.

In the context of QoS assessment, each configuration would be represented by an arm of the bandit. The arm's probability of success is the probability that a simulation run of the given configuration satisfies the QoS requirements. Thompson sampling provides a strategy to identify the optimal configuration wrt. QoS.

However, in situations where it is not necessary to identify the optimal configuration, but rather a configuration that satisfies some QoS requirement with high confidence, standard Thompson sampling tends to put too much effort into optimization, and misses to build confidence in already promising candidate arms. We will outline a solution approach to this problem in the following.


\section{QoS-Aware Thompson Sampling}
\label{sec:QATS}

A basic form of QoS-aware Thompson sampling (QATS) can be realized by determining the probabilities of QoS violation and satisfaction from the arms' belief distributions. In fact, we are interested in the probability $p_\mathrm{v} = P(X \leq q)$ of the true parameter violating the QoS requirement $q \in [0, 1]$. This property can easily be determined from the cumulative density function of the belief distribution.

The probability $p_u = P(X > \hat{p}_i) = 1 - P(X \leq \hat{p}_i)$ that a sampled probability $\hat{p}_i$ from a belief distribution is underestimating the true parameter of an arm is also computable from the belief distribution's cumulative density function.

To solve the exploration vs. exploitation dilemma in a QoS-aware manner, QATS maximizes the odds of underestimation vs. QoS violation. In fact, we prefer large probabilities of underestimation (meaning our belief sample is defensive) while at the same time preferring arms that expose a low probability of QoS requirement violation.
\begin{equation}
	o = \frac{p_\mathrm{u}}{p_\mathrm{v}}
\end{equation}

\begin{algorithm}[H]
	\begin{algorithmic}[1]
		\Procedure {QoS-Aware Thompson Sampling}{}
		\State Sample $\hat{p}_i$ from $\textit{Beta}(s_i + 1, f_i + 1)$ for each arm $i$
		\State Play arm with $\max o_i$ wrt. $\hat{p}_i$ and QoS requirement $q$
		\State Update $s_i$ or $f_i$ according to result
		\EndProcedure
	\end{algorithmic}
	\caption{QoS-aware Thompson sampling.}
	\label{alg:QATS}
\end{algorithm}

\begin{figure}
\centering
\includegraphics[width=0.8\columnwidth]{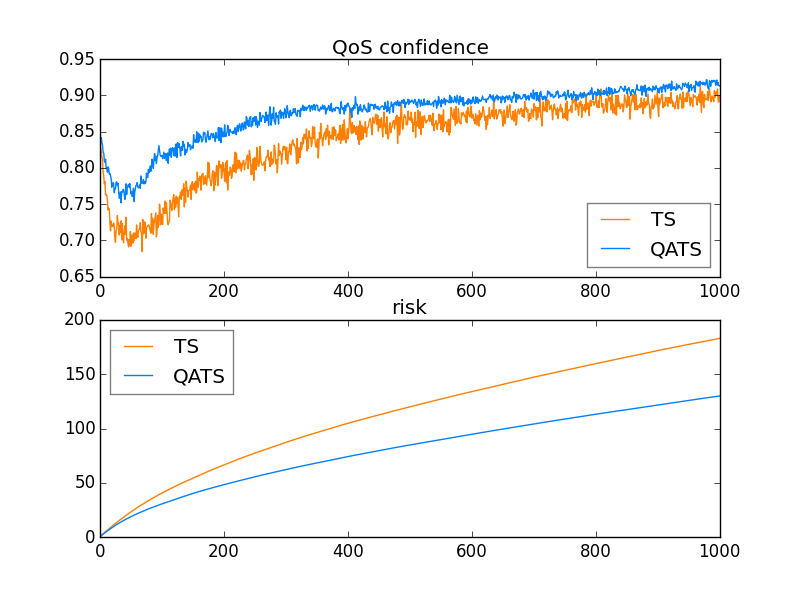}
\caption{QATS vs. TS: QoS confidence in sampled arm (top) and cumulative risk of sampling an arm violating QoS requirements (bottom). 350 runs.}
\label{fig:results}
\end{figure}

QATS is shown in Algorithm \ref{alg:QATS}. We tentatively compared QATS to classic Thompson sampling (TS) in synthetic experiments with promising preliminary results. As an example, consider a four-armed bandit with $p_i \in [0, 0.2]$, instantiated randomly uniform each run. The QoS requirement was set to $q = 0.1$. We evaluated the performance of QATS and TS for 1000 decisions. Figure \ref{fig:results} (top) shows the system's average confidence about QoS satisfaction (i.e. $1 - p_v$) of chosen arms. QATS (blue line) is more confident about QoS satisfaction of chosen arms than TS (orange line). We also measured the cumulative probability of choosing an arm violating the QoS requirement. QATS shows less risk to choose QoS-violating arms. Corresponding results are shown in Figure \ref{fig:results} (bottom).

\section{Conclusion}
Motivated by runtime verification of QoS requirements in self-adaptive and self-organizing systems that are able to reconfigure their structure and behavior in response to runtime data, we proposed QoS-aware Thompson sampling (QATS) for multi-armed bandits. QATS is applicable in settings where QoS satisfaction of an arm has to be ensured with high confidence efficiently, rather than finding the optimal arm while minimizing regret. 

Preliminary experimental results are promising and encourage further research in the field of QoS-aware decision making. It would be interesting to investigate theoretical properties of QoS-aware decision making algorithms. Another direction would be to integrate risk measures (as in financial decision making) into Thompson sampling. See \cite{galichet2013exploration} for a similar approach based on frequentist confidence bounds. Also, QoS-aware decision making could prove useful in sequential decision making, where one decision changes optimality/quality of subsequent decisions.
See \cite{bai2013bayesian} for an application of Thompson sampling in Monte Carlo Tree Search. 
Integration of QoS-awareness into the optimization procedure itself (e.g. the procedure that produces potential system reconfigurations) could allow for even more efficient QoS-aware decision making.

\section*{Acknowledgment}

The authors would like to thank Matthias H\"olzl and Martin Wirsing for insightful discussions.



\bibliographystyle{IEEEtran}
\bibliography{IEEEabrv,literature}
%
%
%

\end{document}